\documentclass[10pt,twocolumn,letterpaper]{article}

\usepackage{cvpr}
\usepackage{times}
\usepackage{epsfig}
\usepackage{graphicx}
\usepackage{amsmath}
\usepackage{amssymb}
\usepackage{multirow}
\usepackage{diagbox}
\usepackage{subfigure} 
\usepackage{authblk}

\usepackage{hyperref}

\cvprfinalcopy 

\begin{document}

\title{Improving Face Detection Performance with 3D-Rendered Synthetic Data}



\author[D. Kopta et al.]
      {Jian Han$^1$, Sezer Karaoglu$^{1,2}$, Hoang-An Le$^1$,
      Theo Gevers$^{1,2}$\\
      $^1$Computer Vision Lab, University of Amsterdam\\
      $^2$3DUniversum\\
      {\tt\small \{j.han, h.a.le, s.karaoglu, th.gevers\}@uva.nl}\\
      }


\renewcommand\Authands{ and }

\maketitle

\begin{abstract}

In this paper, we provide a synthetic data generator methodology with fully controlled, multifaceted variations based on a new 3D face dataset (3DU-Face). We customized synthetic datasets to address specific types of variations (scale, pose, occlusion, blur, etc.), and systematically investigate the influence of different variations on face detection performances. We examine whether and how these factors contribute to better face detection performances. We validate our synthetic data augmentation for different face detectors (Faster RCNN, SSH and HR) on various face datasets (MAFA, UFDD and Wider Face).
\end{abstract}

\section{Introduction}
Face detection is one of the important topics in the field of computer vision. It plays a fundamental role in basically all face related applications. Face detection is the problem of determining the presence of faces in images and their precise locations. Face detection is confronted with different challenges such as variations in scale, pose, expression, occlusion and illumination which all may have a negative influence on the performance of face detection methods. In Table \ref{table1}, we summarized the different characteristics of various face detection benchmarks. From the table, it can be derived that many datasets are limited in representing extreme poses, different scales and heavy occlusions. However, datasets containing face images under a wide variety of imaging conditions are required to develop face detectors which are robust to all variations of image formation process.

Face detectors are designed to address only a limited set of variations in real-world situations. For example, FAN\cite{wang2017face1} uses an attention based structure and data augmentation to cope with facial occlusion. PCN \cite{shi2018real} proposes rotation-invariant face detection in a coarse-to-fine manner by dividing the calibration process into several progressive steps. The HR detector \cite{hu2016finding} combines both features and image pyramids to make the algorithm robustness against extreme face scales. In Table \ref{table2}, we show that different face detectors are designed to cope with different imaging conditions. These face detectors heavily rely on the availability of large-scale annotated datasets. Collecting and annotating real-world datasets with different imaging conditions is tedious, time consuming and in some cases even unfeasible. Furthermore, it is difficult to systematically vary the imaging parameters and to avoid errors during the annotation process. Errors in ground truth may lead to far-reaching impact on training and testing of the networks. Therefore, our contribution is to generate synthetic data, as complementary to real data, to create fully controlled datasets by means of automatic and error-less annotation. To validate our methodology, we train different face detectors on a combination of real data and fully controlled synthetic dataset, to systematically address the imaging variations. Our synthetic images are rendered versions of real 3D faces with changes in viewpoint, scale, illumination, occlusion and background. Hence, the variation of imaging conditions is performed in 3D space. 

Our contributions are (1) we provide a new face dataset (3DU-Face) with a large variety of imaging conditions such as scale, pose, occlusion, blur, etc., (2) large scale experiments to systematically study the impact of data augmentation on the performance of face detection, (3) a comparative study of state-of-the-art face detectors (Faster RCNN, SSH and HR) on different face  benchmarks (MAFA, UFDD and Wider Face). 


\begin{table}
\begin{center}
\begin{tabular}{|l|ccc|}
\hline
\diagbox{Feature}{Datasets}     
&MAFA &UFDD&Wider\\
\hline\hline
landmark occlusion &\checkmark&\checkmark&\checkmark\\
complex background  &  &  &\checkmark\\
extreme pose   &  &&\checkmark \\
extreme scale  &  &\checkmark&\checkmark  \\
heavy occlusion &\checkmark&\checkmark&\checkmark\\
blur &\checkmark &\checkmark&\checkmark  \\
extreme illumination &&\checkmark&\checkmark\\
misleading objects&&\checkmark&\checkmark\\
\hline
\end{tabular}
\end{center}
\caption{Three face detection benchmarks and their characteristics.
}
\label{table1}
\end{table}

\begin{table}
\begin{center}
\begin{tabular}{|l|ccc|}
\hline
\diagbox{Feature}{Detector}&Faster RCNN&SSH&HR\\
\hline\hline
landmark occlusion &\checkmark&\checkmark&\checkmark\\
complex background &\checkmark &\checkmark &\checkmark\\
extreme pose &  & \checkmark &\checkmark\\
extreme scale  &  &\checkmark&\checkmark  \\
heavy occlusion &&&\checkmark\\
blur && &\checkmark\\
extreme illumination &&&\checkmark\\
misleading objects&&&\\
\hline
\end{tabular}
\end{center}
\caption{Three advanced face detectors and their characteristics.
}
\label{table2}
\end{table}
\section{Related Work}
\subsection{Face Detection}
Often face detection is considered as a special case of object detection. Object detectors are, in general, categorized as one-step (e.g. SSD, YOLO) and two-step (e.g. Faster R-CNN) detectors. Two-step detectors mostly use region proposals and classification, while one-step detectors only rely on single feed-forward convolutional networks (without classification). 

Most of the face detectors are designed to address specific variations in real-world scenarios e.g. scale\cite{yang2016multi, yang2017face, zhang2017s, zhu2017cms, hao2017scale, zhang2017s, zhu2018seeing, tang2018pyramidbox}, occlusion\cite{chen2017masquer, ge2017detecting, wang2017face1, sface2018},  pose\cite{shi2018real} or lighting condition\cite{zhou2018hybrid}. Therefore, face detectors are mostly suitable for datasets with corresponding characteristics and may lack generalization power (cross datasets).  
For example, a single face detector may be restricted in handling a wide range of face scales (e.g., 10 px vs. 1000 px tall faces). Therefore, HR uses extremely large receptive fields to locate tiny faces. It also applies multi-scale testing by using an image pyramid to capture extreme scale features\cite{hu2016finding}. To detect occluded faces \cite{ge2017detecting}, a local linear embedding method is used to reduce noise and recover the lost cues. For blurry scenes, Bai et al. \cite{bai2018finding} propose a GAN based super-resolution and refinement network framework. It restores high-resolution faces from blurry ones. For face detection, blurry and low-resolution faces only have few features to extract from. Also, the boundary between the object and background is often difficult to distinguish.


\subsection{Face Synthesis}

Synthetic data is useful to improve the performance in face related applications\cite{masi2016we, osadchy2017genface, abbasnejad2017using, kortylewski2018training}. The generation of synthetic face images can be achieved by face editing methods including shape morphing\cite{blanz1999morphable}, relighting\cite{wang2009face, shu2017neural}, pose normalization\cite{hassner2015effective, zhu2015high, yim2015rotating}, and expression modification\cite{thies2016face2face, pumarola2018ganimation}. Recently, GAN-based methods provide realistic results of facial attribute manipulation \cite{choi2017stargan, shen2017learning, lu2017recent} but they are bounded by the limitations of the training images. Training images merely cover a narrow range of variations, and may cause artifacts during generation. Face synthesis methods are widely used in face recognition tasks\cite{masi2016we, kortylewski2018training}, because such tasks rely on extensive face attribute information. However, in our paper, we focus on generating face images for face detection also considering extreme variations such as large scale and heavy blur. Existing datasets mostly contain subtle face attribute manipulation. Further, existing methods directly manipulate faces in 2D. This may hinder their application in handling extreme imaging conditions. Our synthetic images are rendered versions of real 3D faces. Data generation in 3D space enables the process of including extreme changes in viewpoint, scale, illumination, occlusion and background. 

\subsection{Influence of detection characteristics}

The performance of detectors are affected by both network architecture and object characteristics. Hoiem et al. \cite{hoiem2012diagnosing} provides an extensive analysis on the influence of different variations on different detectors. Another comparison study\cite{huang2017speed} focuses on the trade-off between speed and performance of meta-architecture detectors. 
Karaoglu et al.\cite{karaoglu2016detect2rank} exploits the correlation between different object detectors by means of high-level contextual information. In this paper, we focus on faces and their characteristics. The aim is to study the influence of different data augmentation methods on face detection performance.
\begin{figure*}
\centering
\includegraphics[width=0.8\textwidth]{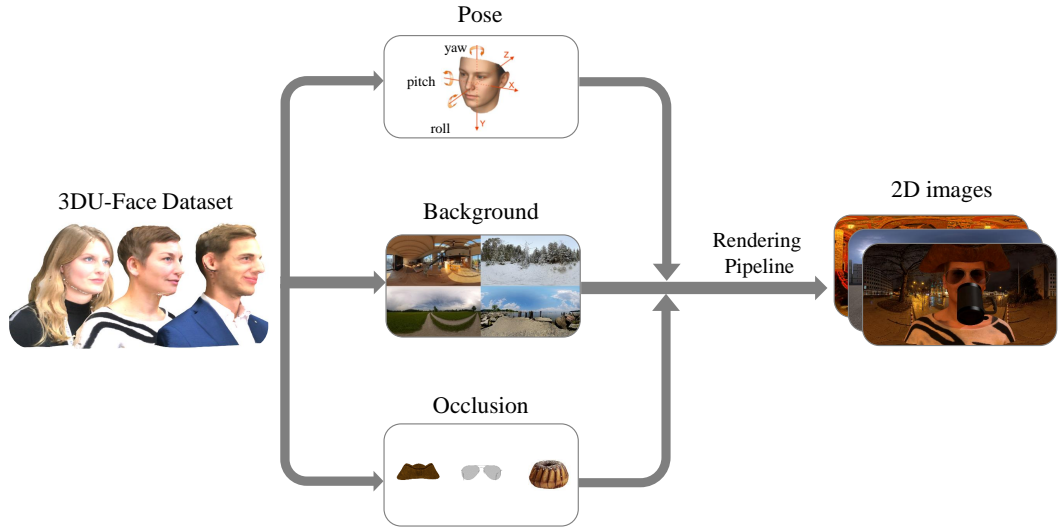}

\caption{An overview of rendering pipeline. It shows the generation of real-synthetic images. First, pose, background and occlusion are manipulated on the original 3D models. Then, the 3D models are converted into 2D images. The figure of the pose variation is taken from \cite{ariz2016novel}.}
\label{fig1}
\end{figure*}
\section{Challenges for Face Detection}
\textbf{Pose} can significantly change the appearance of a face. Extreme poses may result in heavy occlusion or skewed aspect ratio of the face bounding box. Most (deep) face detectors use augmented data by rotating faces over different angles, or jointly estimate the pose\cite{osadchy2007synergistic} to obtain robustness against pose variations\cite{farfade2015multi}. PCN\cite{shi2018real} calibrates the orientation of faces to upright at different stages progressively.

\textbf{Scale} changes may have a negative influence on the performance of face detection. For example, the image features for a 10px face are essentially different than of a 1000px sized face. Combining feature pyramids and multi-scale testing is used to detect faces of extreme scales\cite{hu2016finding}.

\textbf{Context} information can play a crucial role in determining the precise location of faces. Faces in unconstrained settings may be surrounded or occluded by different objects. Round-shaped, background objects may result in false positives. For example, HR\cite{hu2016finding} uses large-scale context information to locate tiny faces. SSH \cite{najibi2017ssh} applies a context module to effectively use background features.  

\textbf{Facial occlusion} may obstruct the presence of valuable information for detection\cite{ge2017detecting}. Facial occlusion can be divided into two different categories: landmark occlusion and heavy occlusion. Landmark occlusion means that only a few landmarks like eyes or mouths are occluded, while most of the face is still visible. In contrast, heavy occlusion means that more than half of the face is missing due to occlusion, image border or extreme pose\cite{chen2017masquer}. Or that a face is occluded by another face.

\textbf{Illumination} changes may substantially influence the appearance of faces. For example, it is difficult to distinguish faces, under extreme lighting conditions, from the background. Zhou et al. \cite{zhou2018hybrid} uses multi-spectrum sensing to detect faces under low lighting conditions.

\textbf{Blur and Low resolution} usually impede face detectors from retrieving available information. For example, images may be distorted in collection, storage, or transmission, leading to degraded quality of images\cite{lowquality}. In some extreme cases, only the outline of the faces can be identified. Bai et al. \cite{bai2018finding} use GANs to refine blurry faces to improve performance. Refinement network or multi-scale testing are feasible solutions to detect blurry or low-resolution faces.

\section{Face Detection: Datasets}
In this section, we discuss a number of well-known face detection benchmarks and their characteristics.

\textbf{MAFA}
is a representative dataset of face images with occlusion. The dataset is mainly composed of occluded samples using different types of occlusions\cite{ge2017detecting}. To cope with the interference of pose, MAFA only includes a narrow range of head poses. MAFA has three types of annotations in the dataset: masked, unmasked and ignored. Faces are extremely blurry or tiny where faces with a side length of less than 32 pixels are labeled as 'Ignored'.

\textbf{UFDD} contains faces in different weather conditions and other challenging variations concerning lens impediments, motion blur and defocus blur\cite{nada2018pushing}. Additionally, it has a collection of distracting images. For the UFDD dataset, the most challenging part is the extreme lighting and blur.

\textbf{Wider Face} is the most challenging benchmark for face detection ~\cite{yang2016wider}. It includes various events (e.g., basketball, football) with a variety of backgrounds. The large number of faces contain extreme poses, exaggerated expressions, heavy occlusion and extreme lighting conditions. The most challenging part of Wider Face is the extreme scale. Wider Face has three categories of difficulty: easy, medium, and hard. The criteria to categorize faces into these different categories are vague. Our Table \ref{table4} shows the basic characteristics of faces, irrespective of invalid faces, in the Wider Face validation partition. 
\setlength{\tabcolsep}{4pt}
\begin{table}[h]
\begin{center}
\begin{tabular}{|l|ccc|}
\hline
Partition & Large & Medium  & Tiny\\
\hline\hline
Height&50-400(96.6\%)&30-50(99\%)&10-30(99\%)\\
Width&20-300(96.3\%)&10-70(99.7\%)&8-20(95\%)\\
Number &7211 &6108 &18636\\
\hline
\end{tabular}
\end{center}
\caption{Face scale information of the validation set in Wider Face. We distinguish three face categories based on height and width. Proportion information represents the percentage of faces that fits within the scale interval.}
\label{table4}
\end{table}
\setlength{\tabcolsep}{1.4pt}

\subsection{Face Detectors} 
In this section, we outline the different face detection algorithms used for comparison.

\textbf{Faster RCNN} is one of the mostly used object detectors in the literature. It is not designed to be robust against challenging variations \cite{ren2015faster} for face detection. 

\textbf{SSH} is an extremely fast one-step face detector. It is designed to be scale invariant \cite{najibi2017ssh}. To accelerate the inference process, it removes selectively a number of parameters from the structure. This strategy has a negative influence on the detector's performance. SSH requires additional multi-scale processing to detect faces with extreme scales.

\textbf{HR} face detector performs well on tiny faces by using wide-range contextual information and using testing on multiple resolutions \cite{hu2016finding}. Its architecture resembles RPN \cite{ren2015faster} and uses both feature pyramids and image pyramids. However, HR face detector trained on Wider Face is extremely sensitive to tiny, round objects in the background. HR heavily relies on contextual information to locate faces. For faces with limited information (e.g., heavily occluded, extremely small or blurry), complex background may hinder precise detection. 
\begin{figure}[!htbp]
\centering
\includegraphics[width=0.4\textwidth]{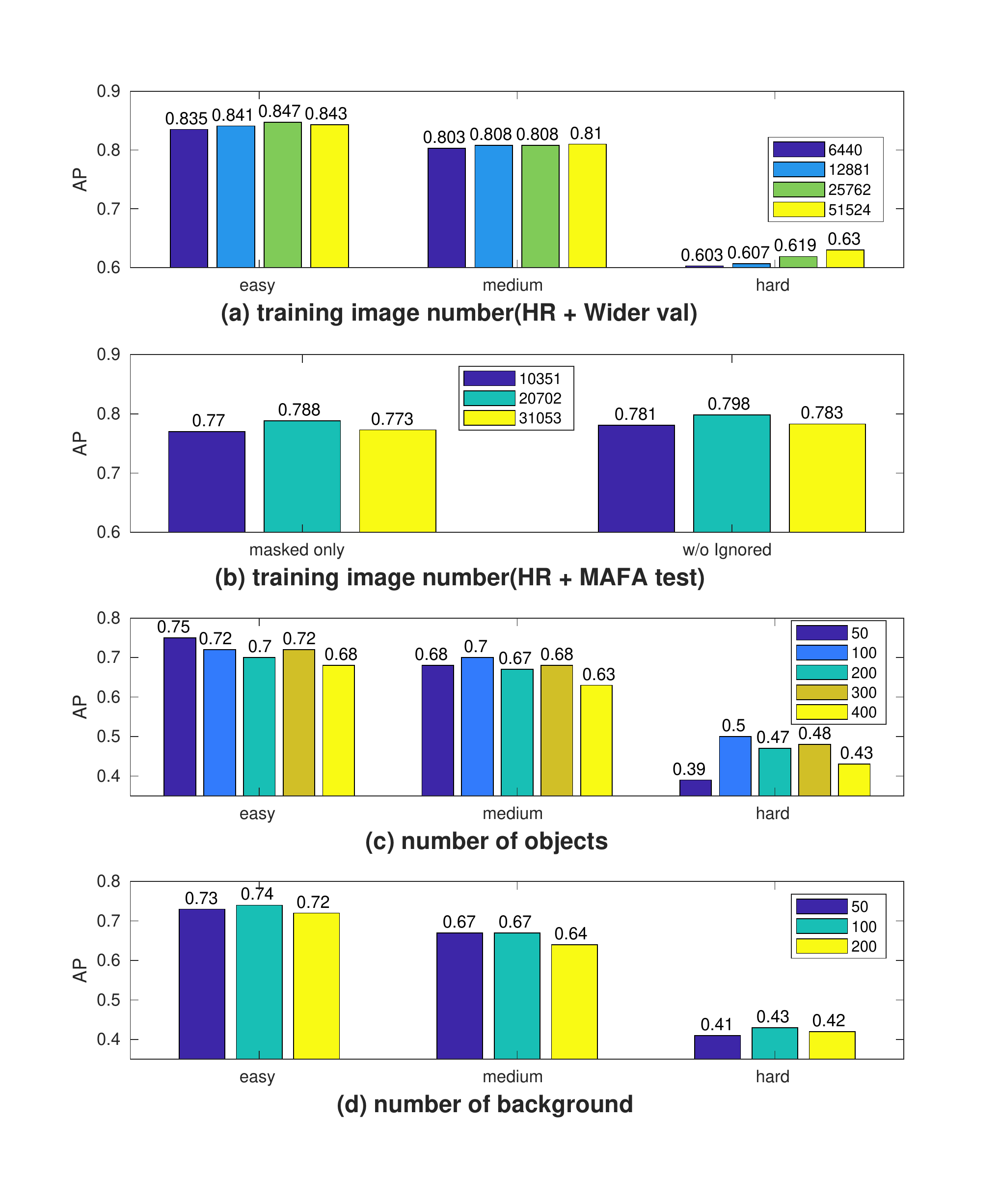}

\caption{Performance comparison on the basic settings of our training data. It includes the effect of training data size on both (a) Wider Face validation set and (b) MAFA test set. (c) and (d) respectively show the effects of objects and background on Wider Face validation with HR.}
\label{fig3}
\end{figure}
\section{Data Augmentation}
Data augmentation is based on our new 3D face dataset (3DU-Face). It contains 700 3D face mesh models with high-resolution texture of 435 different individuals. Some people may have multiple recordings taken at different times and imaging conditions. Most of the 3D models are taken in the wild under uncontrolled conditions. The 3D models contain 50 facial landmarks annotated by human experts. 

Using the 3D models, the 2D images (projections) are rendered. The rendering pipeline is built in Blender 2.78. To change the viewpoint, the model is rotated over different Euler angles. The camera is kept in the same position. The parameters of pitch, yaw and roll are selected randomly using different ranges. For face scale variation, we change the distance between camera and the face models within a fixed range. The ground truth for face detection is generated from the 3D landmarks. The bounding box is generated to tightly encompass the forehead, chin, and cheek. The size of the faces is larger than $10\times8$ pixels.

In this paper, data is created that include face occlusion. Current approaches focus on (1) cropping face images or (2) using GAN to generate. However, both approaches have drawbacks. Cropping will reduce the information of face and may not provide sufficient generalization in case of real occlusion samples. GAN-based methods are able to generate subtle face attributes. However, for more global image/face changes, GAN may generate blurry results and artifacts. Therefore, in this paper, our aim is to generate face images in 3D space.  We randomly add different 3D objects like sunglasses, hats, and helmets in the 3D space before rendering. All objects are placed at a selective locations to simulate landmark occlusion. To simulate occlusion, face regions are divided into three different face parts: head, eye and mouth. More than 1000 different combinations of occlusion are generated for each model. 

\section{Results}

Experiments are conducted to systematically study the influence of data augmentation on the performance of face detection. We compare three face detection methods on the following face detection benchmarks: MAFA, UFDD and Wider Face. The face detection methods are Faster RCNN\cite{ren2015faster}, SSH\cite{najibi2017ssh}  and HR\cite{hu2016finding}. 
\begin{figure}[t]
\centering
\includegraphics[width=0.4\textwidth]{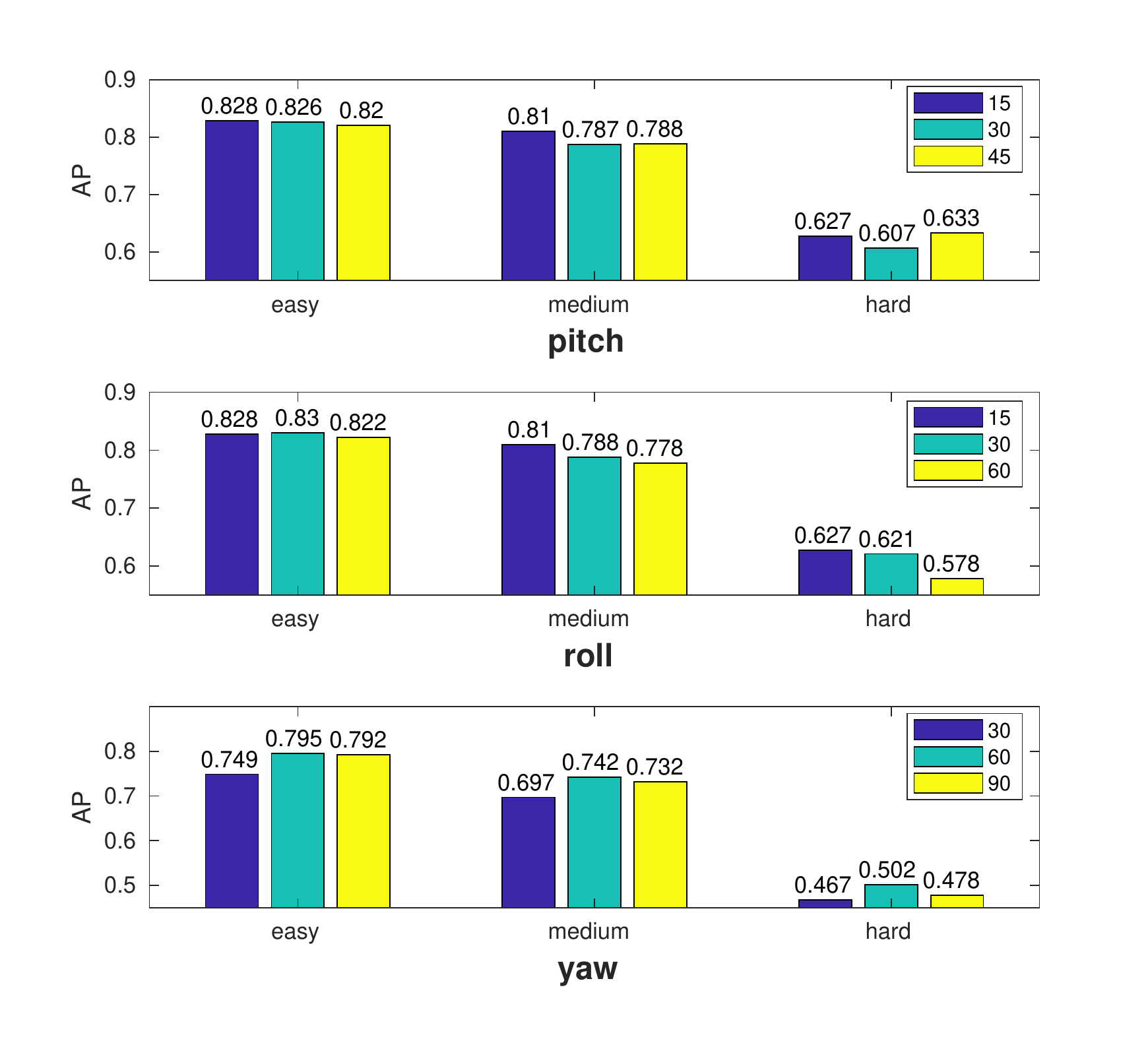}
\caption{Performance comparison on pose variation. Across pitch, yaw, and roll, different colors represent the rotated degree, as labeled at the top right (e.g., "15" means "-15" to "15".) }
\label{fig4}
\end{figure}
\subsection{Implementation Details}

\subsubsection{Settings of rendering process}
This section is to demonstrate the basic settings of our rendering process . The 3D models are not changed in terms of shape or texture in experiments. We choose 100 fixed 3D models as experimental subjects and set the default for pitch, roll and yaw randomly in ranges of (-15, 15), (-15, 15), (-60, 60), respectively. For each face model, the rotation origin is the center of its landmarks, irrespective of invalid landmarks. All the face models are aligned by using landmarks with an anchor model. The anchor model is aligned to the global axis in Blender. Within a fixed range (from 1 to 48 faces), we randomize the number of faces in each image. The distance between each model and camera is randomly selected within the range of 0 to 20 meters. The size of the faces is larger than $10\times8$ pixels. 50 HDR images (no humans) are taken from Shape Net \cite{shapenet2015} and used as backgrounds. These images provide environment lighting and background variations for the synthetic images. The background dataset includes both indoor and outdoor scenes. Back face culling is applied to avoid artifacts in the rendered results.

\subsubsection{Settings of the face detectors}
In the following experiments, we test our data augmentation methods with different face detectors on three benchmarks. We first train the face detectors on synthetic data and test them on real data.   We also validate the methods on a subset of real data for the training part. Every dataset has its own domain. There are many different parameters in our rendering pipeline. It is unlikely to find the optimal setting for a specific real dataset. However, after comparing the performance on real data with different rendering parameters, we attain suitable and effective configurations for testing. We use the augmented synthetic data to improve the performance for real data. Different face detectors have their own approaches to augment data, like flipping, cropping, or transforming images. For fair comparison, we keep their original operations and hyper-parameters. For SSH and HR, we deploy their algorithms on one single GPU\cite{bal2016medium}. For Faster RCNN, we use the implementation from \cite{frcnn_pytorch}.

\subsection{The influence of data augmentation} \label{DataAugmentation}
In this section, we study the influence of various data augmentation formats. For this purpose, HR is considered. We do not employ Faster RCNN or SSH for extensive analysis but we rather use them to test performance of data augmentation. Faster RCNN is a generic detector without multi-scale testing, of which the performance may not reflect all the changes in variations. SSH is designed to be an scale-invariant one-step detector. For all experiments, we study one variation at the time. The other variations are kept the same.

\textbf{The influence of basic training data settings}:
We change the number of images to test the influence of training dataset size. As shown in Figure \ref{fig3}.(a), the increase of synthetic images continuously improves the performance on hard, but not easy and medium, levels. This is because more than half of the faces, for the hard level, are tiny faces. In the training process, large faces generate much more positive samples than tiny faces. Hence, larger training samples are more useful for tiny faces. Simply increasing the number of synthetic images may lead to over-fitting. Because most of faces from the Wider Face dataset contain extreme scales, a test is performed on MAFA to examine our conclusion about the (training) dataset size in Figure \ref{fig3}.(b). It shows that the performance on both occluded faces and full datasets (including occluded and unoccluded faces) is saturated after adding more training images. Moreover, background is crucial. HDR images provide higher resolution and less sharper results than real images. HDR images have their own bias in respect to other images; the increase of the number of HDR images does not improve the performance constantly (see Figure \ref{fig3}.(d)). In Figure \ref{fig3}.(c), we study the influence of the number of 3D objects. It shows that the number of 3D models does not strongly influence the performance. This may be because the face detection task does not heavily rely on the attribute or identity information of the faces. Also, our 3D models may have their own bias (e.g. annotation bias, mesh corruption or scanner noise).

\begin{figure}[t]
\centering
\includegraphics[width=0.5\textwidth]{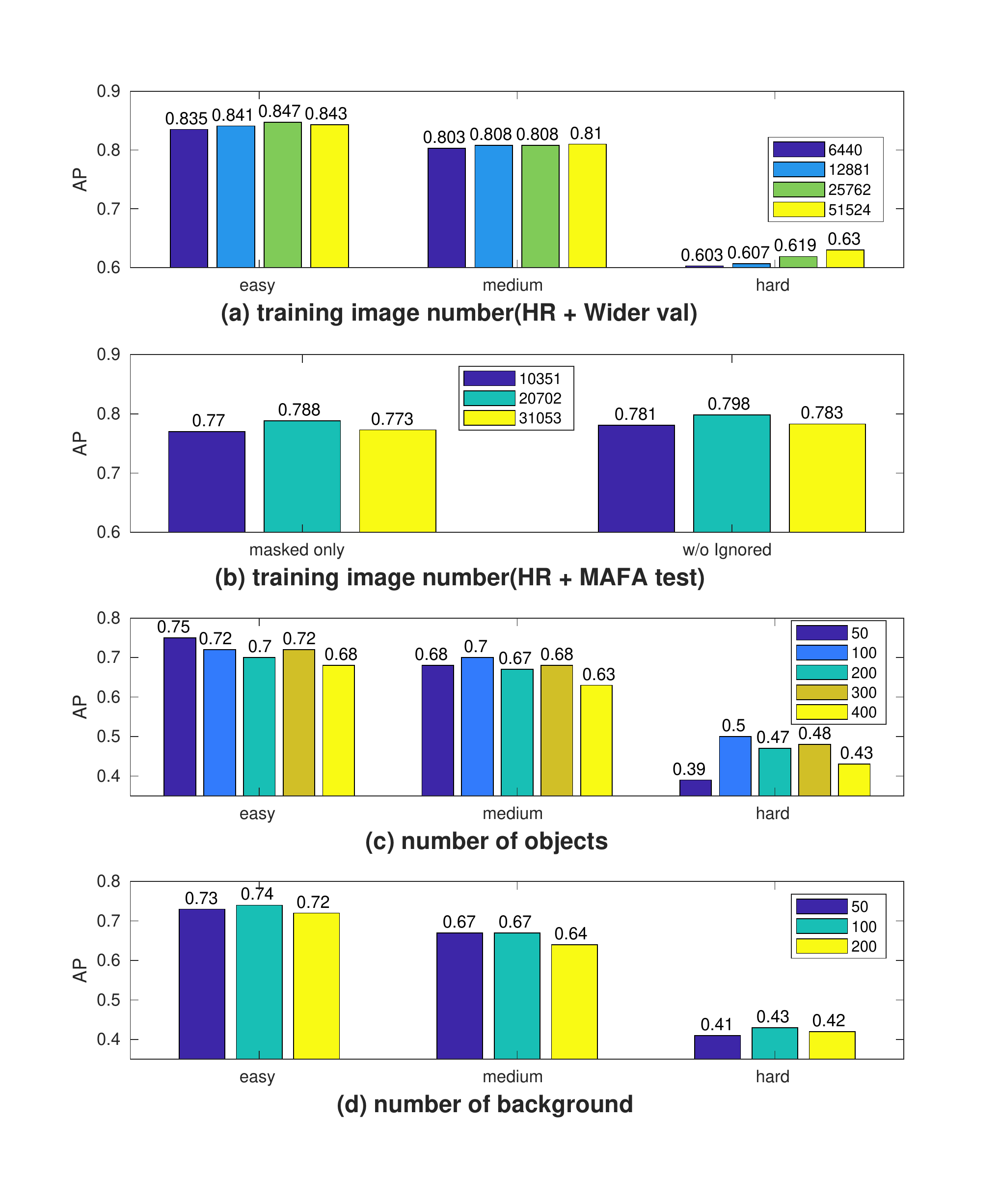}
\caption{Performance comparison on the basic settings of our render pipeline. It includes the effect of training data size on both (a) Wider Face validation set and (b) MAFA test set. (c) and (d) respectively show the effects of objects and background on Wider Face validation with HR.}
\label{fig3}
\end{figure}

\begin{figure}[t]
\centering
\includegraphics[width=0.4\textwidth]{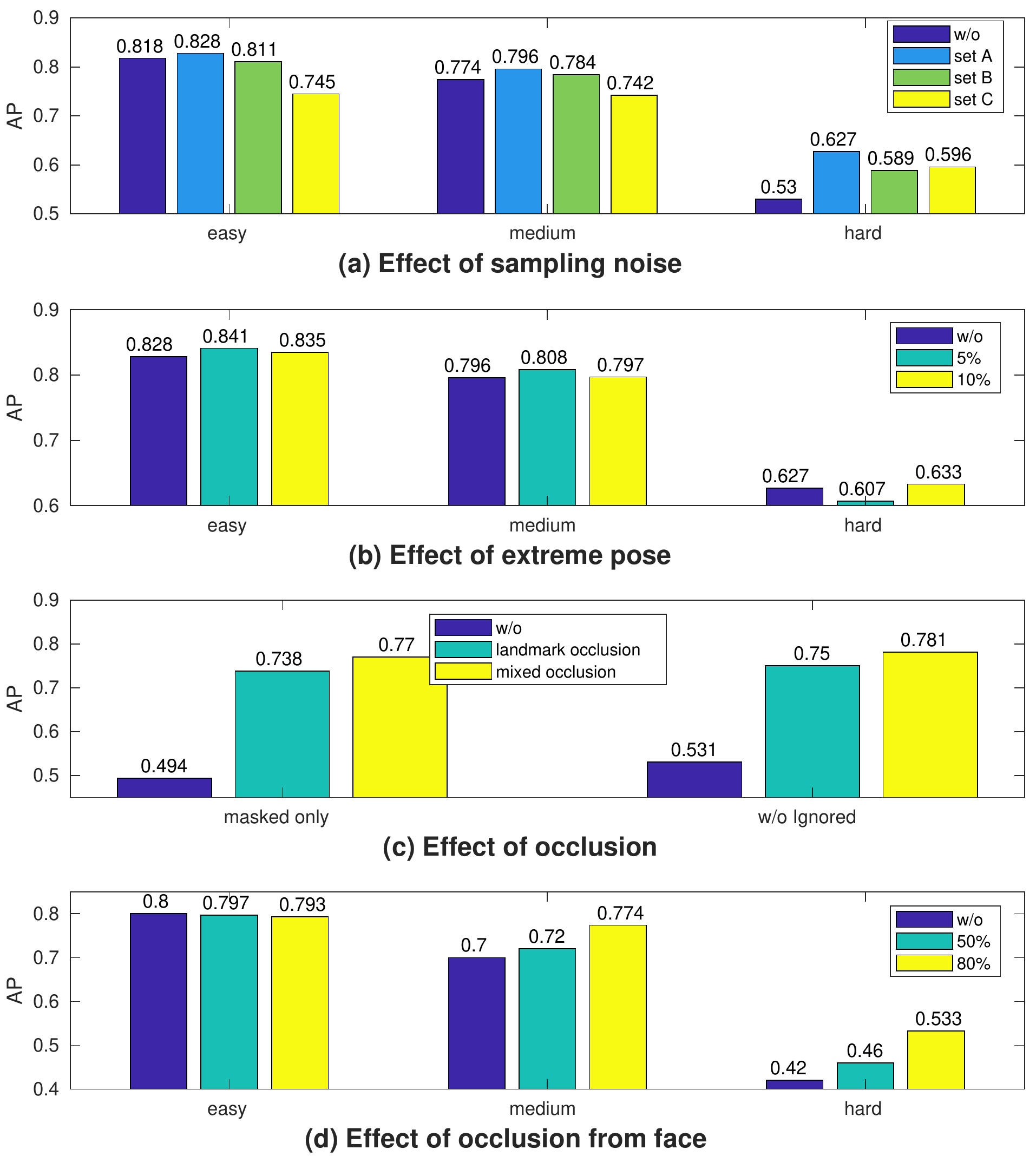}
\caption{Performance comparison on other variations. Only (c) tests on MAFA test set, while the remaining are on Wider Face validation set. (a) shows the results of adding different noise levels from the down-sampling or up-sampling process; (b) shows the results after adding small-portions of extreme poses to the training dataset; (c) shows the results of adding different types of occlusion; (d) shows the results of adding occlusion from other faces.}
\label{fig5}
\end{figure}

\textbf{The influence of pose}: 
To investigate the effect of head pose, we render faces with different ranges of pitch, roll and yaw, and test them on the Wider Face validation dataset. As shown in Figure \ref{fig4}, for pitch, roll or yaw, the minimal range gives better performance than others. The reason is that the majority of faces in the dataset do not contain extreme poses. Also, the face detectors need sufficient data to learn the representation of faces. Then, the different portions of extreme orientations are added to the training dataset, see Figure \ref{fig5}.(b). A small range of extreme pose boosts the performance of large and medium faces because most of the tiny faces (hard level) are extremely blurry and pose-agnostic. For the HR detector, most of the tiny faces are detected by multi-scale testing.  Features from generic faces are more useful for detecting tiny faces.

\textbf{The influence of occlusion}:
We test two different types of face occlusion. The first type of occlusion is from other objects in the scene. MAFA focuses on the occlusion of face images. We test our augmentation methods for this first type of occlusion on the MAFA test set for three different occlusion settings: the baseline condition with no extra occlusion, landmark occlusion, and mixed occlusion (both landmark and heavy occlusion). For all occluded cases, none of the 2D faces is occluded by parts of the other 3D face models. As shown in Figure \ref{fig5}.(c), the performance on the MAFA test set improves drastically after adding occlusion in the synthetic training dataset. HR becomes more robust after training on synthetic faces with landmark and heavy occlusions. Some occlusion examples from our synthetic data are shown in Figure \ref{fig1}.

The second type of occlusion is from other faces or human body parts. We choose the Wider Face validation set, to study how this second type of occlusion in our synthetic training data influences the performance of HR. This is because most images in the Wider Face are acquired using unconstrained settings. Many of the images are group pictures, and each image may have hundreds of tiny faces. We set a threshold for overlapping faces. This is to avoid other large faces to cover the tiny faces. As shown in Figure \ref{fig5}.(d), after adding occlusion from other faces in the training dataset, the performance of HR for hard level in Wider Face validation set improves substantially. The results on the first type of occlusion demonstrates the effectiveness of our occlusion augmentation. Our synthetic data provides suitable noise to simulate the patterns of occlusion. As for the second kind of occlusion, the samples in training dataset are needed for detectors to learn to distinguish the boundary between different faces.


\textbf{The influence of noise}: 
Each benchmark has its own type of configuration. For the Wider Face dataset, the original high-resolution images are downloaded using a search engine and resized to a predetermined width of 1024 pixel. This process introduces noise caused down-sampling or up-sampling. Therefore, we first render images with multiple resolutions (as Set A, B and C as below), and then re-size them to one fixed resolution (1024$\times$768). Set A includes multiple high-resolution images (4096, 3072, 2048). Set B has multiple high- and low-resolution images (4096, 3072, 2048, 512, 256, 128). And Set C has multiple low-resolution images(512, 256, 128). We demonstrate the influence of noise for the different difficulty levels of Wider Face in Figure \ref{fig5}.(a). The performance for all the difficulty levels of Wider Face has been improved, especially for tiny faces. Set A achieves the best performance on Wider Face. This is because sampling process is also changing the size of the original faces in the rendered images. The tiny faces for real data are resized from large faces as our operation in Set A. As for Set B and C, a part of the large and medium faces are resized from the original tiny faces in the rendered images. 

\subsection{Performance comparison on synthetic data}
In Table \ref{table5}, we show the performance comparison about three synthetic data sets on Wider Face validation set. These three synthetic datasets $s_1, s_2, s_3$  are combined with real data to improve detection performance for UFDD\cite{nada2018pushing} and Wider Face respectively in Section \ref{UFDD} and \ref{Wider Face}.
\setlength{\tabcolsep}{4pt}
\begin{table}[h]
\begin{center}
\begin{tabular}{|c|ccc|}
\hline
Set & Easy & Medium  & Hard\\
\hline\hline
$s_1$&0.795&0.742  &0.502\\
$s_2$&0.818&0.774  &0.53\\
$s_3$&0.828&0.796  &0.627\\
\hline
\end{tabular}
\end{center}
\caption{Average precision from HR\cite{hu2016finding} trained on different sets of synthetic data. These three sets are combined with real data to improve detectors' performance. Different settings: $s_1$ is our basic settings for rendering with light occlusion. $s_2$ combines $s_1$ with extra occlusion from other faces in the render process. $s_3$ adds additional blurry results from down-sampled high resolution images into $s_2$.}
\label{table5}
\end{table}
\setlength{\tabcolsep}{1.4pt}
\subsection{Improving face detector performance by data augmentation}
In this section, we study how to use our synthetic data to improve the performance of the face detectors (i.e., Faster RCNN, SSH and HR) on real datasets. We train on a combination of Wider Face and synthetic data and then test on MAFA, UFDD and Wider Face. Visualization of our detection results are shown in Figure \ref{fig9}. 
\begin{figure}[t]
\centering
\includegraphics[width=0.4\textwidth]{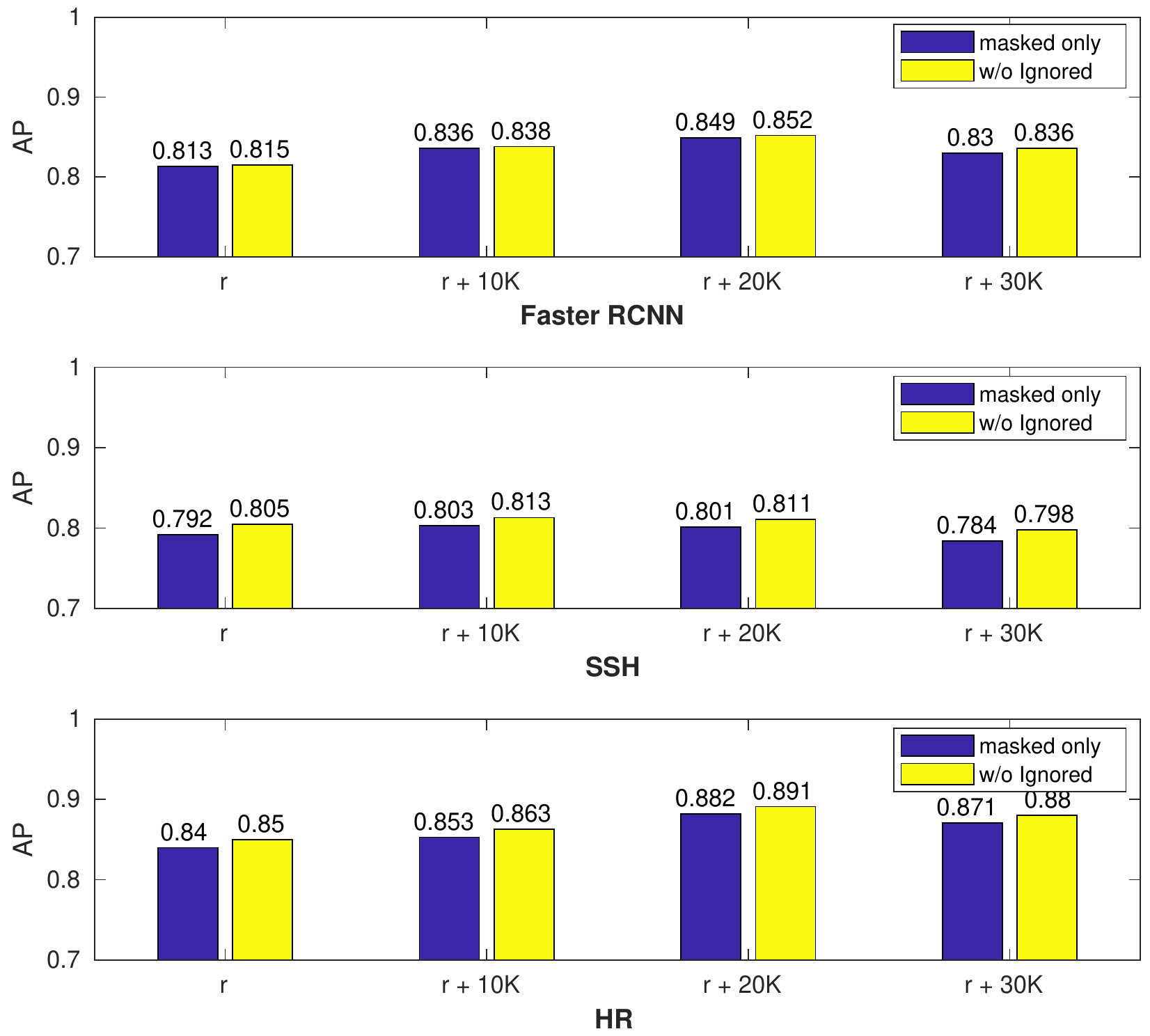}

\caption{Performance comparison on  different sizes of synthetic data in training on MAFA test set with different detectors. }
\label{fig6}
\end{figure}

\begin{figure}[t]
\centering
\includegraphics[width=0.4\textwidth]{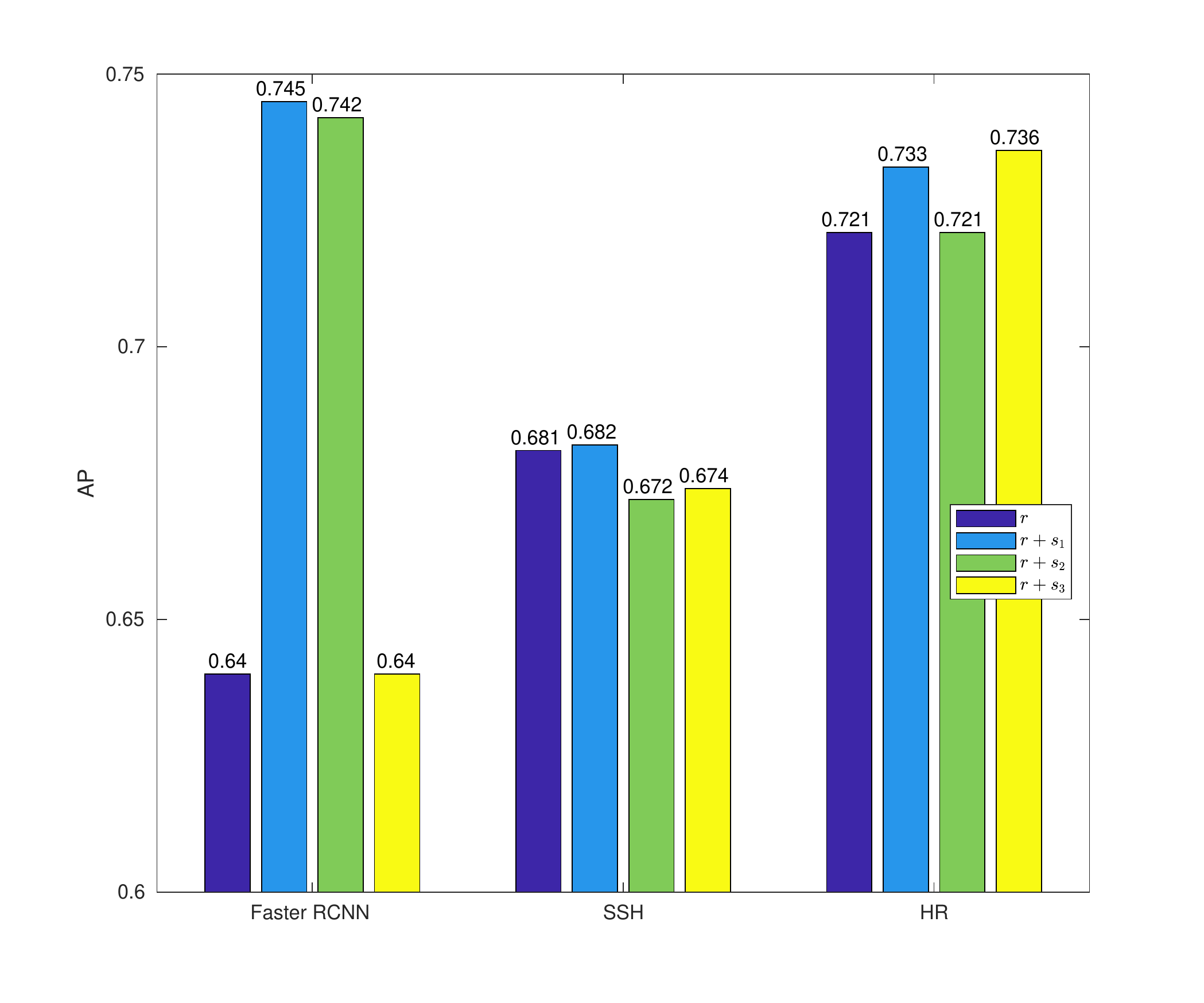}

\caption{Performance comparison on different data augmentations on UFDD test set with different detectors.{the table is not visible and not clear.}}
\label{fig7}
\end{figure}

\begin{figure}[!htbp]
\centering
\includegraphics[width=0.4\textwidth]{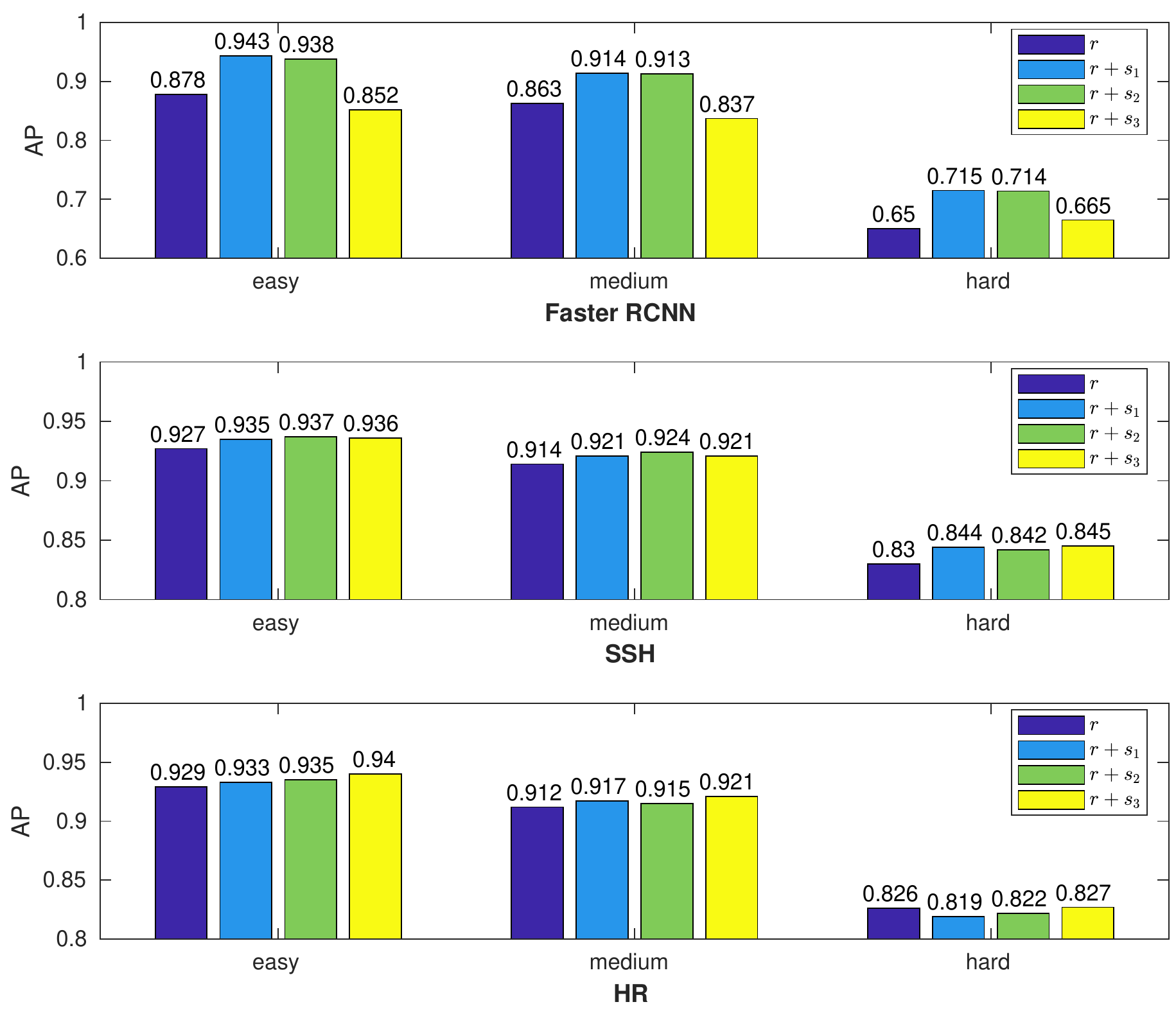}
\caption{Performance comparison on different data augmentations on Wider Face validation set with different detectors.}
\label{fig8}
\end{figure}

\begin{figure*}[h]
\centering
\includegraphics[width=0.8\textwidth]{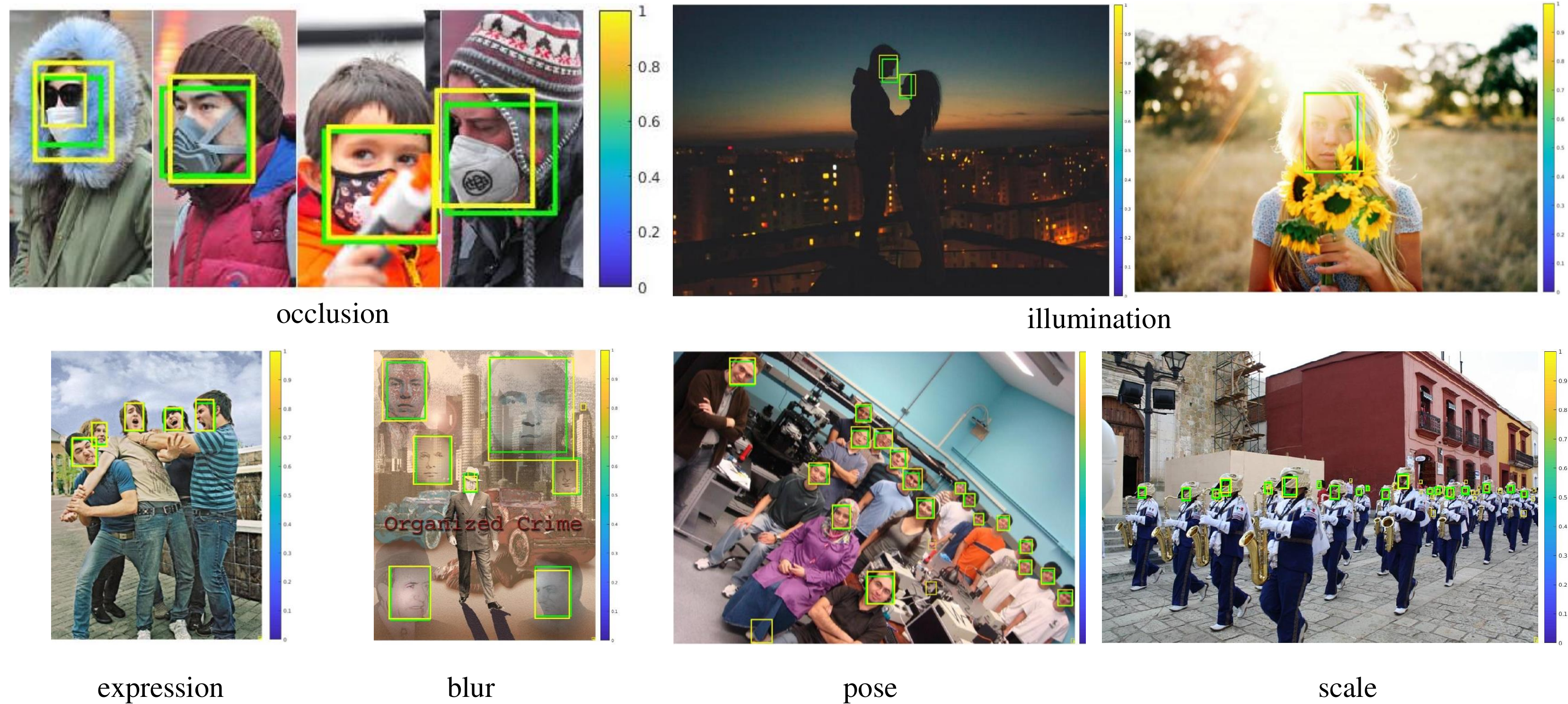}
\caption{Qualitative results on different variations of real dataset. We visualize examples of each variation.}
\label{fig9}
\end{figure*}

\begin{figure*}[h] 
\centering
\includegraphics[width=0.8\textwidth]{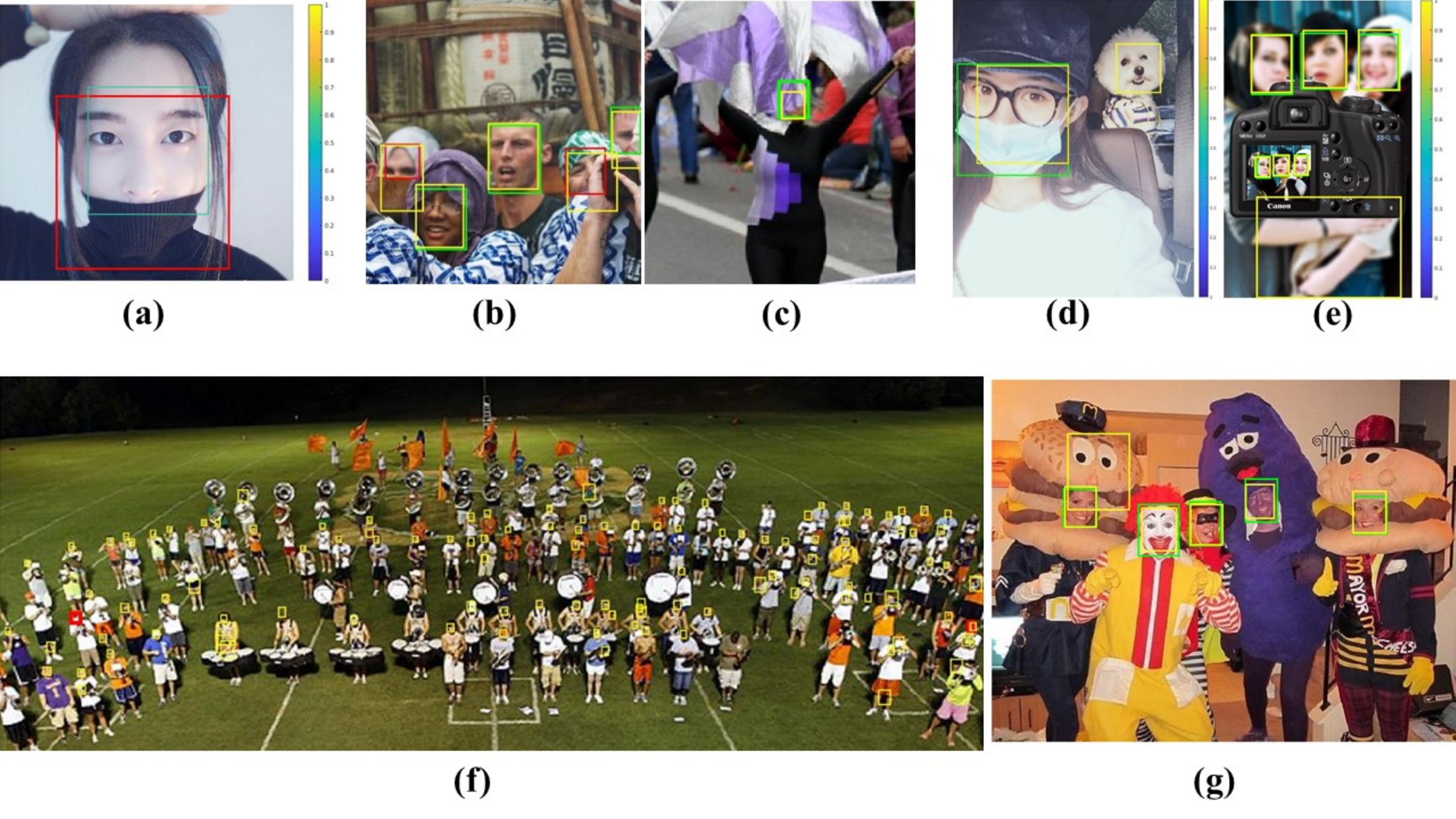}
\caption{False positive examples derived from our detection results. In both Figure \ref{fig9} and Figure \ref{fig10}, the green bounding boxes are the ground truth boxes; the bounding boxes with other colors are predictions with different confidence intervals; and the red bounding boxes are false negative examples. (a) is a square annotation example from the MAFA test set. (b) and (c) are occlusion annotations of, respectively, UFDD and Wider Face. (f) only has three annotations for ground truth but actually has more unlabeled faces.(d), (e) and (g) are a number of examples of the real dataset.}
\label{fig10}
\end{figure*}
\subsubsection{MAFA}
We only use mixed face occlusion (that is landmark and heavy occlusion as in \ref{DataAugmentation} for data augmentation for MAFA. This is because during the collection and annotation process, MAFA has constraints on the different types of variations. The synthetic images for data augmentation follow the same setting of MAFA training set. Our hypothesis is that the training data size will have an effect on the performance. As shown in Figure \ref{fig6}, the performance of different detectors improves with the increase of the number of synthetic images. However, after adding more and more data, the performance saturates and drops. The reason could be that there is a domain gap between the synthetic and real data. Our synthetic data may not be as complex as real data. Also, our synthetic data inherits biases (e.g. annotation bias or mesh corruption) from the original (real) 3D models. Therefore, there is a inverted U-shape relationship between the increase of synthetic data in training and performance. 

\subsubsection{UFDD} \label{UFDD}
In this section, our data augmentation is used with the Wider Face only. This is because (1) the training dataset of UFDD is not made public, and (2) UFDD itself is trained on Wider Face \cite{nada2018pushing}. Three synthetic datasets $s_1, s_2, s_3$ are combined with real data to improve detectors' performance. Different settings: $s_1$ is our basic settings for rendering with light occlusion. $s_2$ combines $s_1$ with extra occlusion from other faces in the render process. $s_3$ adds additional blurry results from down-sampled high resolution images into $s_2$. The performance of data augmentation is shown in  Table \ref{table5}. The influence of our data augmentation is shown in Figure \ref{fig7}. After merging synthetic data and real data together, the performance of Faster RCNN, which is trained on real data, improves significantly on $r + s_1$ and $r + s_2$. Given that Faster RCNN is not trained for different scales, the noise of $s_3$ impedes its performance. As for SSH, its architecture and parameters heavily rely on Wider Face. For UFDD, its performance becomes saturated after being trained on real data. After adding synthetic data, its performance is even worse than Faster RCNN. Faces in UFDD are not very challenging to HR; the performance therefore only changes slightly after using our data augmentation.

\subsubsection{Wider Face} \label{Wider Face}
We still use the same setting $s_1, s_2, s_3$ as in \ref{UFDD} to perform data augmentation for Wider Face. The performance comparison is shown in Figure \ref{fig8}. Faster RCNN is a generic object detector without multi-scale testing. Hence, it is supposed to generate fewer predictions than HR and SSH. After we add synthetic data, the performance substantially improves for all levels. The performance of HR and SSH nearly saturate after being trained on real data. 

\subsection{Analysis}
\subsubsection{Analysis of synthetic data}
The advantage of synthetic data is that the variations in dataset can be fully controlled. Although, there is always a domain gap between synthetic data and real data, this paper shows that synthetic augmentation can provide large-scale datasets with annotations conveniently and precisely. Our results show the applicability of synthetic data as an alternative to real data.

\subsubsection{Analysis of false positives}

False positives are the primary factor that decrease the performance. In Figure \ref{fig10}, we plot a number of false positive examples. There are two major sources of false positive in our detection results. 

The first source is annotation. Different datasets have their own annotation process. This process may have a negative influence on our predictions. For example, the annotation of occluded faces in Wider Face is based on the region of the entire face (Figure \ref{fig10}.(c)). In comparison, UFDD often annotates the visible part of occluded faces ((Figure \ref{fig10}.(b)). MAFA uses square annotation, which may contain background information of surrounding faces (Figure \ref{fig10}.(a)). Moreover, human annotators may not be able to annotate all the tiny and blurry faces in the background, see Figure \ref{fig10}.(f).

The second source of false positives is due to confusing objects, such as round-shaped objects and human body parts. Because real data has much more diverse and complex background than synthetic data. Our synthetic images are rendered from 3D face models. Most of these 3D face models only depict the upper part of the human body. Therefore, our rendering results are inherently unrepresentative for other human body parts. As a result, other human body parts (see Figure \ref{fig10}.(e)) and accessories can be a source of false positives. 

\subsubsection{Analysis of face detectors}
Based on our detection results, we analyze the characteristics of three detectors respectively (1) Faster RCNN is an object, instead of face, detector. It does not adjust its settings of anchors for the face detection benchmarks. And it has fewer predictions without multi-scale testing. Despite that, our synthetic data augmentation substantially improves its performance on multiple challenging datasets. (2) SSH is a face-targeted detector. However, our synthetic data augmentation is not able to outperform Faster RCNN in most of the detection tasks except for the hard level of Wider Face. Designed to cope with scale, SSH results in worse performance when  other variations are encountered. Detectors have a trade-off between speed and performance\cite{huang2017speed}. SSH pursues fast speed in inference process such that its light weight architecture is equipped to handle other variations. (3) Although the HR face detector already has excellent performance for different kinds of variations, our synthetic data still improves its performance. However, HR has a drawback that is extremely sensitive to small round-shape objects given its tiny-face-targeted architecture. HR generates more false positives than other detectors. It restricts the generalization on regular faces.
\section{Conclusion}

In this paper, we provide a synthetic data generator methodology with fully controlled, multifaceted variations based on a new 3D face dataset (3DU-Face). We customized synthetic datasets to address specific types of variations (scale, pose, occlusion, blur, etc.), and systematically investigate the influence of different variations on face detection performances. We validate our synthetic data augmentation for different face detectors (Faster RCNN, SSH and HR) on various face datasets (MAFA, UFDD and Wider Face).

{\small
\bibliographystyle{ieee}
\bibliography{egpaper_final}
}

\end{document}